\documentclass[conference]{IEEEtran}
\IEEEoverridecommandlockouts
\usepackage{cite}
\usepackage{amsmath,amssymb,amsfonts}
\usepackage{algorithmic}
\usepackage{graphicx}
\usepackage{textcomp}
\usepackage{xcolor, colortbl}

\def\BibTeX{{\rm B\kern-.05em{\sc i\kern-.025em b}\kern-.08em
    T\kern-.1667em\lower.7ex\hbox{E}\kern-.125emX}}
\begin{document}
\newcommand{\norm}[1]{\left\lVert#1\right\rVert}

\title{Unsupervised Prediction of Negative Health Events Ahead of Time\\
}

\author{\IEEEauthorblockN{1\textsuperscript{st} Anahita Hosseini}
\IEEEauthorblockA{\textit{Department of Computer Science} \\
\textit{University of California Los Angeles}\\
Los Angeles, US \\
anahosseini@ucla.edu}
\and
\IEEEauthorblockN{2\textsuperscript{nd} Majid Sarrafzadeh}
\IEEEauthorblockA{\textit{Department of Computer Science} \\
\textit{University of California Los Angeles}\\
Los Angeles, US  \\
majid@cs.ucla.edu}
}

\maketitle

\begin{abstract}
The emergence of continuous health monitoring and the availability of an enormous amount of time series data has provided a great opportunity for the advancement of personal health tracking. In recent years, unsupervised learning methods have drawn special attention of researchers to tackle the sparse annotation of health data and real-time detection of anomalies has been a central problem of interest. However, one problem that has not been well addressed before is the early prediction of forthcoming negative health events. Early signs of an event can introduce subtle and gradual changes in the health signal prior to its onset, detection of which can be invaluable in effective prevention. In this study, we first demonstrate our observations on the shortcoming of widely adopted anomaly detection methods in uncovering the changes prior to a negative health event. We then propose a framework which relies on online clustering of signal segment representations which are automatically learned by a specially designed LSTM auto-encoder.  We show the effectiveness of our approach by predicting Bradycardia events in infants using MIT-PICS dataset 1.3 minutes ahead of time with 68\% AUC score on average, using no label supervision. Results of our study can indicate the viability of our approach in the early detection of health events in other applications as well.

\end{abstract}


\section{Introduction}
    The increasing prevalence of continuous health monitoring through bedside care and wireless health sensors offers great potential for gaining insights into health state of individuals. However, the majority of the rich collected data remains unlabeled, mainly due to uncontrolled and real-time collection setting and tediousness of offline labeling by domain experts. Therefore, conventional supervised data analysis models which heavily rely on annotations become disadvantageous. 
    
    Unsupervised deep representation learning models, such as auto-encoders\cite{autoencoder}, have recently gained considerable attention in learning informative realization of data, including images\cite{ae-image} and text \cite{ae-nlp}. When it comes to time series analysis, one important application of these models has been anomaly detection, which primarily focuses on recognition of an abrupt change in time-series normal behavior. Such a change is shown to increase the reconstruction error on a model that is trained to generate the normal signal, as the model cannot reconstruct anomalous data points accurately\cite{anomaly-general}. The success of deep anomaly detection in time series has been recently expanded to health care, especially in the analysis of ECG signals\cite{anomaly-ecg}.
    
\begin{figure}[htbp]
\centerline{\includegraphics[width=0.4\textwidth]{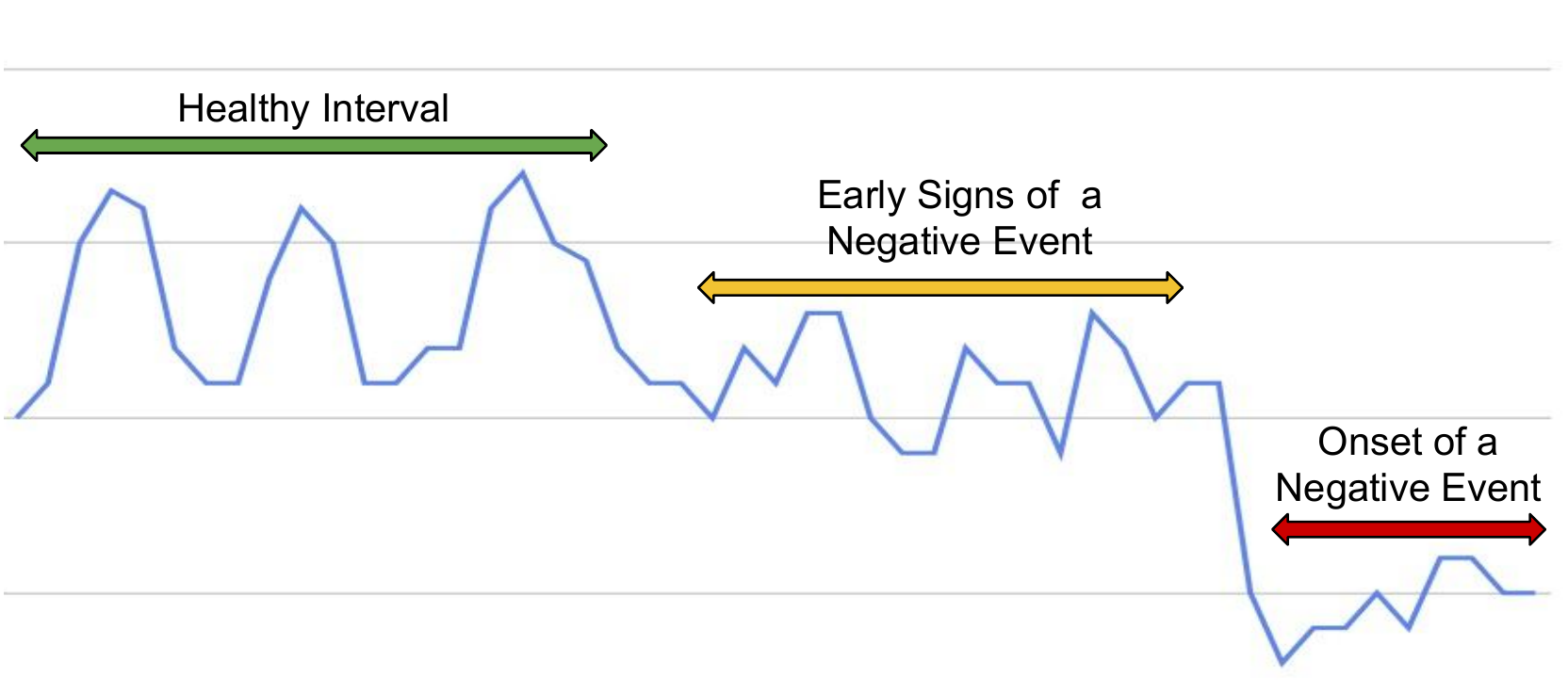}}
\caption{Negative health events may be detectable ahead of time by careful analysis of early signs, which in real-world can be gradual and subtle. }  
\label{fig-first}
\end{figure}

In spite of advances in real-time detection of anomalies, there has been a missing focus on the early prediction of forthcoming negative health events, which can be possible with the unsupervised analysis of health signals in intervals before the event onset. Physiological and environmental changes detected through health sensors can be an early sign for the onset of a negative health event in the near future (Fig.~\ref{fig-first}) and a number of studies have already validated this hypothesis in applications such as prediction of an asthma attack in children \cite{breatheJ} and Bradycardia in infants \cite{bradycaria}.

% Therefore, the missing focus on the analysis of health signals in episodes before the onset of an event needs to be addressed.
    
    In this study, we propose an unsupervised approach based on deep sequential auto-encoders and online clustering of the internal representations to address the aforementioned problem. We use the PICS dataset \cite{physionet} which is recently made available to predict the onset of Bradycardia heart events in infants. We show that the widely adopted anomaly detection methods relying on the increase of reconstruction error perform poorly on distinguishing the more subtle and complex changes of signal behavior in pre-event episodes. Instead, we analyze the clusters formed by the representation of signal segments from an auto-encoder using Denstream \cite{denstream}, an online and noise-tolerant clustering method. In the design of the auto-encoder, we employ an LSTM encoder-decoder based architecture \cite{lstm-ae} alongside wavelet transform of the signal \cite{wavelet} to capture temporal features of time and frequency domain. Furthermore, we use unit-ball regularization of the learned representations \cite{Nokia} to optimize the results of our clustering phase.
    
    In short, to the best of our knowledge, our study is first to address the problem of future negative event prediction using unsupervised models and propose a framework that its prediction capability is validated on the early detection of Bradycardia events in infants, 

% In short, in this study, we emphasize on the importance of addressing the problem of forthcoming negative event prediction and propose a generalizable approach and validate its capabilities by evaluating it on early prediction of Bradycardia events in infants and in comparison to existing methods. 
% This observation can be due to the fact that reconstruction error can optimally capture sudden and sharp behavior changes in signal rather than gradual ones.[] 
\section{Related Work}
With the rise of unsupervised deep learning models especially auto-encoders\cite{autoencoder} and their great performance in other domains such as image recognition \cite{ae-image}, their application has recently emerged in wireless health for detection of anomalies in health signals such as ECG signals. \cite{anomaly-ecg, anomaly-ecg2} are among studies that employed auto-encoders on ECG to distinguish anomalous parts from the healthy ones. For this aim, the reconstruction error from the auto-encoder that is trained on normal data is tracked to find sudden jumps, motivated by the idea that such a model cannot reconstruct anomalous intervals of data accurately. Auto-encoders have successfully replaced prior approaches such as classifiers \cite{old-anomaly-survey} which require large annotated datasets, alongside statistical clustering models \cite{anomaly-clustering}, and future value predictor models \cite{anomaly-old-futureValue} that both are not easily generalizable to other applications. 

LSTM auto-encoders \cite{lstm-ae} were later introduced in learning representations of videos and improved feature extraction by capturing temporal features of the signal. They were later used in time series analysis as well\cite{anomaly-ecg2}. Moreover, Two recent studies have shown improved performance of auto-encoders in more complex anomaly detection settings by utilizing the encoded representation from auto-encoders in offline clustering of anomalies \cite{Nokia} or detection of signal change point by comparing neighbor segment representations \cite{changepoint-ae}. Although these studies follow different goals, we employ their finding in this study in building our model.

Prediction of Bradycardia in infants using the PICS dataset was approached before by publishers of the dataset with statistical methods \cite{bradycaria}. They specifically used a point process analysis and tried to capture the differences in variance and mean of signal segments before a Bradycardia event. Although this study proves the feasibility and achieves reasonable accuracy, their approach is supervised, hand-engineered, and heavily relies on the observance of multiple onsets of Bradycardia events in each infant, which is not always possible in the real-world setting. This is while our approach focuses on the straightforward collection of normal signals from individuals and the detection of changes in an unsupervised and automatic manner. 

%Furthermore, they noted some differences in low frequency components of signal in episodes before an onset.

\section{Methodology}

Learning an unsupervised representation of health signals can be used to distinguish intervals of data that may lead to a negative event. In this section, we review the representation learning and online clustering approaches used for this aim.
\subsection{Sequential Representation Learning}
In time series data, temporal features carry important information. Therefore, we employ a variant of LSTM encoder-decoder architecture similar to \cite{lstm-ae} to encode signal segments into informative representations.
\begin{figure}[htbp]
\centerline{\includegraphics[width=0.35\textwidth]{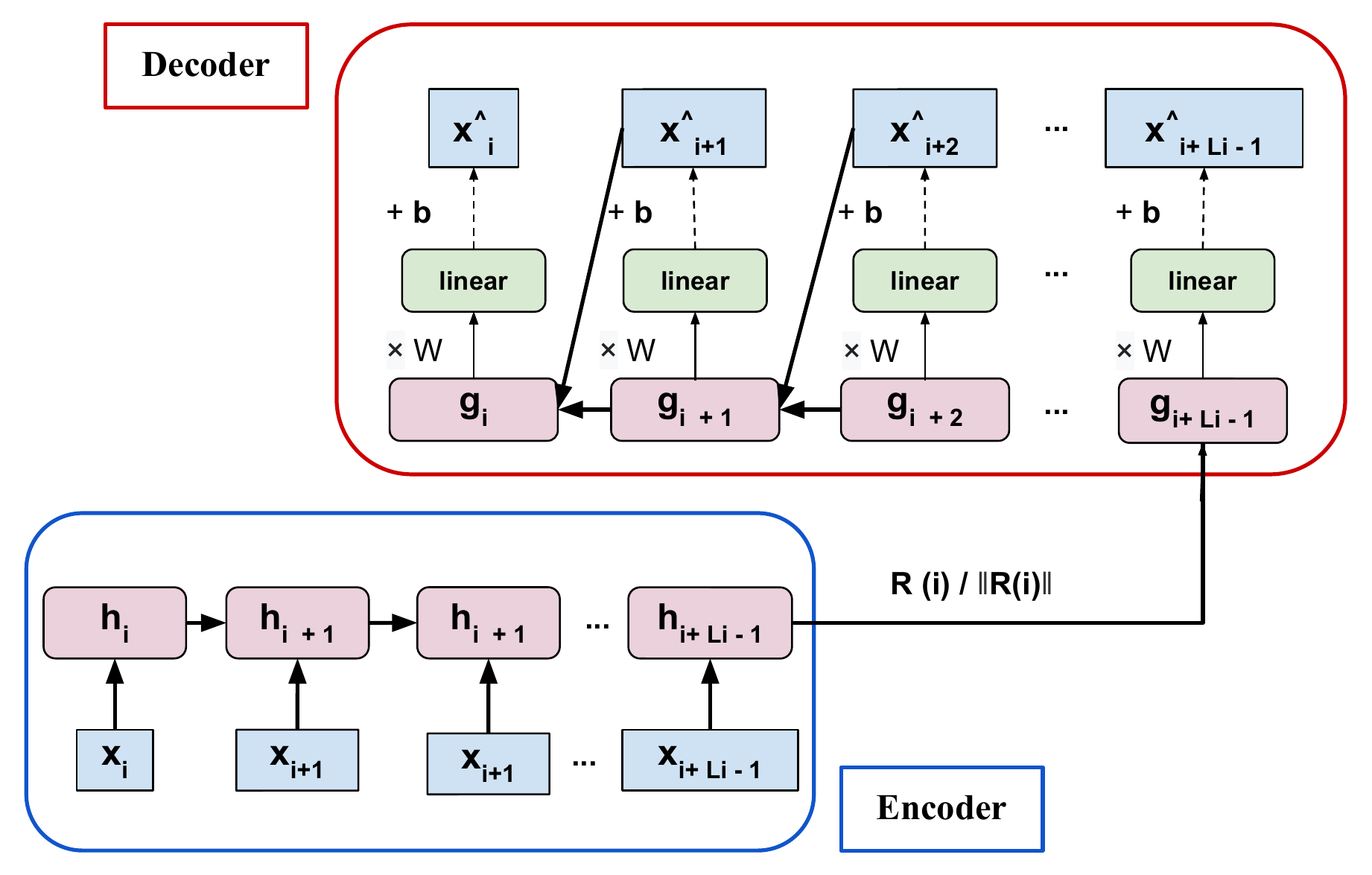}}
\caption{Design of the LSTM auto-emcoder with unit-ball regularization.}
\label{fig-lstm}
\end{figure}

Given a time series $T$ segmented into fixed or variable length windows denoted as $w_1, w_2, w_3, ..., w_n \in T$ where each $w_i$ is itself a list of readings of length $l_i$: 
\begin{equation}
w_i = (x_j, x_{j+1},..., x_{j + l_i -1})  
\end{equation}

The model first embeds $w_i$ into a fixed length representation by feeding it into an encoder module, an LSTM based recurrent neural network (RNN) with $l_i$ cells. The hidden state of the last ($l_i$th) cell can be considered as a compact and informative representation of $w_i$, which we call it $R(i)$. To make these representations more suitable for comparison using distance-based metrics and remove the impact of representation length, we apply $l2$ normalization on $R(i)$ in the training phase and before feeding to the decoder as suggested by \cite{Nokia}.
\begin{equation}
R^*(i) = \frac{R(i)}{\norm {R(i)}}
\label{eq:norm}
\end{equation}

The decoder module that tries to reconstructs window $w_i$ from $R^*(i)$ is also an LSTM RNN with a linear layer on the output gate. It uses $R^*(i)$ as the initial state to the first cell. Also, the output of each cell like $\hat{x_j}$ in the decoder is used as an input to the next cell and also represents the prediction of one reading in $w_i$. We follow findings of prior studies on improved optimization of encoder-decoder architectures \cite{lstm-ae} and predict each window $w_i$ in reverse order. Fig.~\ref{fig-first} depicts the design of the encoder and decoder modules.

Considering decoder module as a function $D$, $D(R^*(i))$ denotes the output of the model for window $w_i$. When reconstruction error between $D(R^*(i))$ and input $w_i$ is used as the objective function and both modules are jointly trained on normal intervals of data, the model learns to embed representative features of a normal input window $w_i$ into $R(i)$. Therefore, the objective function can be written as:

\begin{equation}
L = \frac{1}{J} \sum_{j \in J} (reverse(w_{j}) - D(R^*(i))) + \norm{W}
\label{eq:loss}
\end{equation}
Where, the last term denotes normalization of linear layer weights.

% As it is represented, the output of the encoder module is normalized before being fed into the decoder. Moreover, 
\subsection{Online Clustering}\label{section-cluster}
As discussed, when auto-encoder is trained to reconstruct normal windows of time series, the encoder module, in turn, learns to extract representative features of a normal window. It is hypothesized that deviations from the norm in the signal will reflect in these features. It is important to note that representation of normal windows can come from multiple clusters and recognizing them is important for the detection of an abnormal window. Also, noisy abnormal deviations should be ignored. To reach these goals, we employ Denstream \cite{denstream}.

Denstream, an online and noise-tolerant clustering approach, tries to find close groups of data points as core micro-clusters and by marking those that do not reach a density threshold as outliers, it tackles the noise in data. Real clusters of data points (that we use) are formed by the connection of neighbor core micro-clusters at each point in time. 

Having a trained encoder module, we feed representation of training windows to Denstream to extract main clusters of normal windows, denoted as $C$. In the test time, as the representation of incoming signal windows are extracted and fed into Denstream in real-time, an increased appearance of clusters other than $C$ (abnormal clusters) in a short time is considered as an abnormality and possible event onset. It is worth mentioning that as Denstream removes sparse outlier windows, the abnormal clusters detected are dense enough to show a real change in the signal. In particular, we consider the last $k$ received windows ( calling them "confidence windows" ) and keep track of the number of windows that join an abnormal cluster. A threshold on this score, 50 \% in this study, is used to generate an alarm for an even onset. 

\section{Experiments}
In this section, we review the used dataset, details of pre-processing, experiment setup, and finally our results.
\subsection{Dataset and Data Pre-processing}
Preterm Infant Cardio-Respiratory Signals Database(PICS) \cite{physionet} contains 20 to 70 hours of ECG recordings of 10 infants with multiple onsets of Bradycardia episodes, in which infant's heart rate stays below 100 bpm for at least two beats. 

In this study, we employ the heart rate variability signal, generated by extracting the time difference between R-peaks in ECG. Furthermore, we take Morlet Continuous Wavelet Transform (CWT) \cite{wavelet} of this signal in the low-frequency band (0.01-0.15 HZ) to train our models. Both techniques have been shown to be a powerful tool in the detection of heart events \cite{wavelet-tut}. The pre-processed signal is then segmented into 64-heartbeat windows and fed to the model. The hyper-parameters of Denstream, $\epsilon$, min-neighbor, and decay are set to 0.01, 2, and 0 respectively. The first third of the dataset for each infant is used for model training and the rest for the evaluation. Furthermore, to select only normal intervals of data for the training phase, a 3 and 6-minute margin before and after a Bradycardia event has been disregarded. 
\subsection{Evaluation}
We evaluate our model by scoring the rate of true-positive alarms (recall), true negatives decisions (specificity), AUC, and earliest prediction time on true-positive alarms. An alarm is considered as true positive if a negative event happens within a 3-minute time span. Moreover, 6 minutes after each onset is disregarded in evaluation to ensure the effects of the last event has passed. Therefore, a true negative happens when no alarm is generated from 6 minutes after to 3 minutes before two consecutive events. The used time ranges are borrowed from in the initial study on this database \cite{bradycaria}.

\subsection{Results}
\begin{figure}[t!]
\centerline{\includegraphics[width=0.35\textwidth]{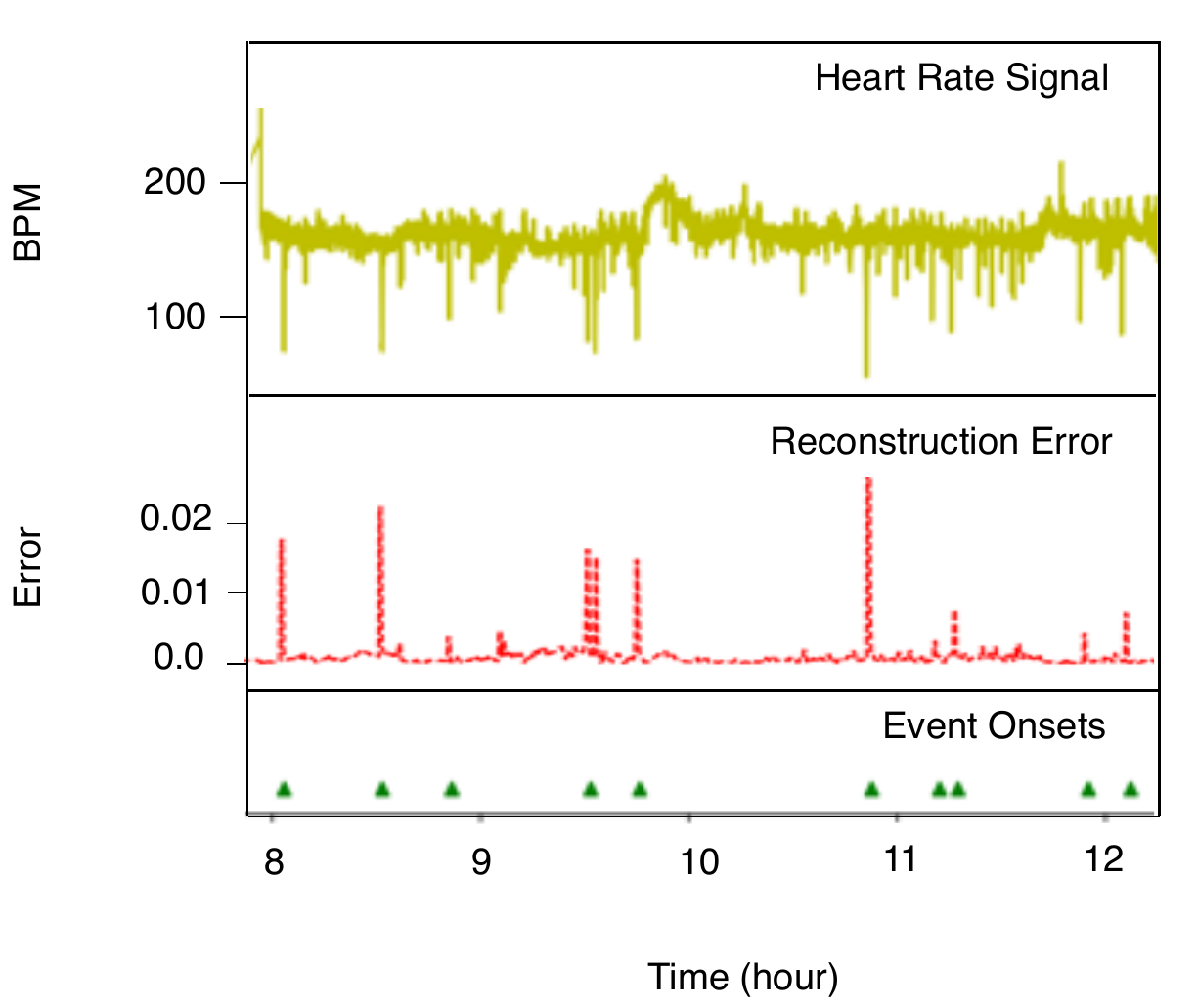}}
\caption{Correlation of heart rate variability signal and reconstruction error in a time span containing multiple onsets of Bradycardia related to infant 7.}
\label{fig-recons}
\end{figure}

\begin{figure}[!b]
\centerline{\includegraphics[width=0.4\textwidth]{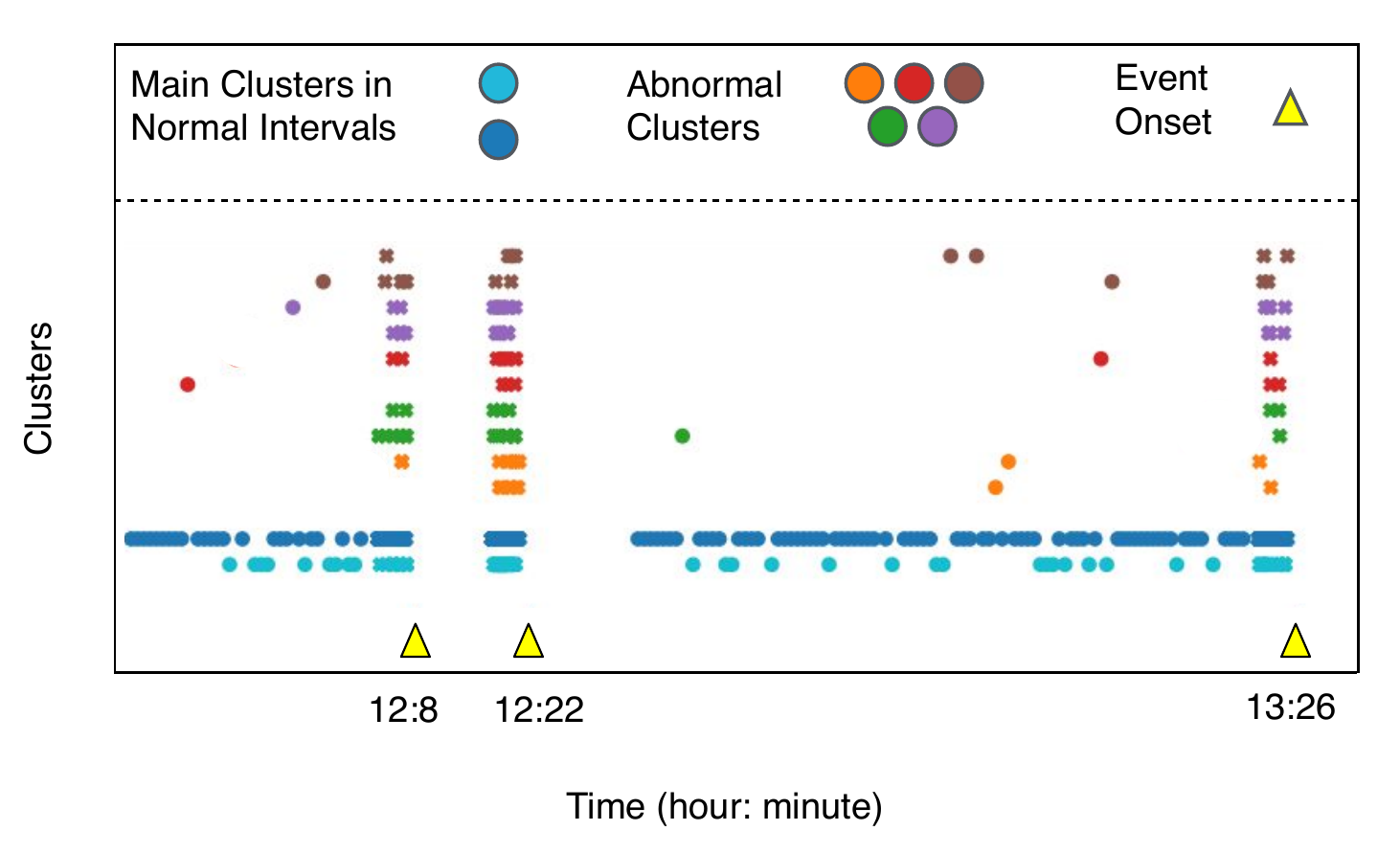}}
\caption{Example of online clustering in an interval of data from infant 7. }
\label{fig-clusters}
\end{figure}
As tracking reconstruction error is used as a common approach in anomaly detection \cite{anomaly-ae}, we share results of evaluating its performance in prediction of future negative events.
As it is noticeable in Fig.~\ref{fig-recons}, reconstruction error performs well in capturing visible and sharp changes in the signal. However, as discussed in \cite{bradycaria} and observable here, the variability in heart rate, which translates into similar variability in the reconstruction error, does not show a simple pattern before a Bradycardia onset. Therefore, although reconstruction error achieves good performance in unsupervised detection of sudden changes, it can perform poorly for prediction of forthcoming events, mainly due to more complex nature of this task. This experiment validates our approach in employing deeper features of signal for uncovering the hidden changes before a negative event .

We next evaluate our proposed model qualitatively and quantitatively. Fig.~\ref{fig-clusters} shows a qualitative view of online clustering results before three events onsets of infant 7. Each cluster is depicted by a unique color and level in the y-axis and those appearing in 3-minute time span before a negative event are shown with a cross mark. We can observe the trend of change in clusters as we move in time (on the x-axis) and process new incoming windows. As the figure suggests,  the two blue clusters that appear in most of the times are related to the normal behavior of data. More importantly, we can observe the sudden appearance of numerous abnormal clusters in the 3 minute time span before each event, showing a powerful sign of an onset. Furthermore, it is noticeable that abnormal clusters appear far more sparsely in normal intervals. Confidence windows introduced in section ~\ref{section-cluster} help in tuning the sensitivity of our model to these appearances.

\begin{figure}[!t]
\centerline{\includegraphics[width=0.3\textwidth]{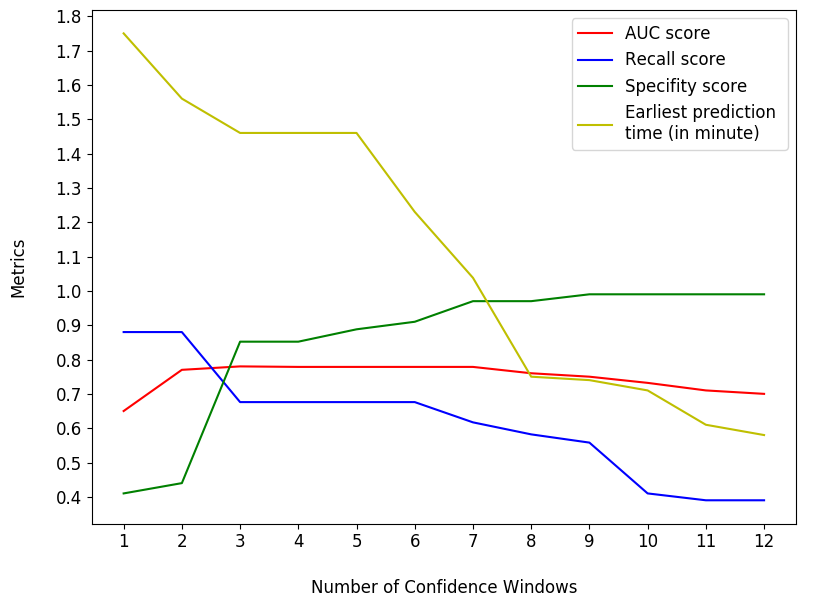}}
\caption{Comparison of evaluation metrics using different confidence windows.}
\label{fig-res-k}
\end{figure}

Results of the next experiment, depicted in Fig.~\ref{fig-res-k}, is used to analyze the impact of confidence window size ($k$) on the performance of our model. In general, as we increase $k$,  the earliest time to prediction and recall decreases while the specificity increases, meaning that false alarms decrease in cost of losing detection of some events. If we observe closer, for $k > 2$ to $k < 7$ we can see a stable performance. This is because having $k > 2$ ensures that a single appearance of an abnormal window does not generate an alarm and $k < 7$ corresponds to around 2 minutes before an event where main changes happen.  We can also observe that AUC is pretty stable as this metric is not dependent on our cut-off threshold (50 \% abnormal observations in a confidence window) and mainly measures how well our model can assign distinguishable scores to positive and negative labels. 

Table ~\ref{fig-res-k} contains AUC score and earliest time of prediction of our model with confidence window of 5 for all infants. The achieved results are competitive with ones from the prior study \cite{bradycaria} (mean AUC of 0.79) when considering definite advantages of our unsupervised approach in the healthcare setting that labels are generally missing.

\begin{table}[!t]
\caption{Results of our model for all the subjects}
\begin{center}
\begin{tabular}{|c|c|c|c|c|c|}
\hline
 \cellcolor{gray!40} Infant Id & \cellcolor{gray!40} 1 & \cellcolor{gray!40} 2 &  \cellcolor{gray!40} 3  & \cellcolor{gray!40} 4   & \cellcolor{gray!40} 5   \\ \hline  
\cellcolor{gray!40} AUC score   & 0.71 &0.65 & 0.61 & 0.68 & 0.63  \\ \hline

\cellcolor{gray!40} Earliest prediction & 0.87 & 1.31 & 1.24& 1.42 & 1.28 \\ \hline \hline

\cellcolor{gray!40} Infant Id &\cellcolor{gray!40}  6   & \cellcolor{gray!40} 7   &\cellcolor{gray!40} 8   & \cellcolor{gray!40} 9  & \cellcolor{gray!40} 10\\ \hline
\cellcolor{gray!40} AUC score  & 0.69 & 0.77 & 0.67 & 0.63 & 0.72 \\ \hline

\cellcolor{gray!40} Earliest prediction & 1.34 & 1.46 & 1.42 & 1.33 & 1.23\\ \hline
\end{tabular}
\label{tab1-res}
\end{center}
\end{table}
\section{Conclusion}
In this study, we approached the problem of early negative health event prediction. We first demonstrated poor performance of common anomaly detection models in addressing this problem and then proposed an unsupervised framework using LSTM auto-encoders and Denstream online clustering. We evaluated performance of our model qualitatively and quantitatively and validated its capabilities for addressing prediction of Bradycardia event in infants using MIT-PICS dataset, achieving average 68 \% AUC score and 1.3 minute early prediction time.
\bibliography{root}
\bibliographystyle{IEEEtran}

% \bibitem{b1} G. Eason, B. Noble, and I. N. Sneddon, ``On certain integrals of Lipschitz-Hankel type involving products of Bessel functions,'' Phil. Trans. Roy. Soc. London, vol. A247, pp. 529--551, April 1955.
% \bibitem{b2} J. Clerk Maxwell, A Treatise on Electricity and Magnetism, 3rd ed., vol. 2. Oxford: Clarendon, 1892, pp.68--73.
% \bibitem{b3} I. S. Jacobs and C. P. Bean, ``Fine particles, thin films and exchange anisotropy,'' in Magnetism, vol. III, G. T. Rado and H. Suhl, Eds. New York: Academic, 1963, pp. 271--350.
% \bibitem{b4} K. Elissa, ``Title of paper if known,'' unpublished.
% \bibitem{b5} R. Nicole, ``Title of paper with only first word capitalized,'' J. Name Stand. Abbrev., in press.
% \bibitem{b6} Y. Yorozu, M. Hirano, K. Oka, and Y. Tagawa, ``Electron spectroscopy studies on magneto-optical media and plastic substrate interface,'' IEEE Transl. J. Magn. Japan, vol. 2, pp. 740--741, August 1987 [Digests 9th Annual Conf. Magnetics Japan, p. 301, 1982].
% \bibitem{b7} M. Young, The Technical Writer's Handbook. Mill Valley, CA: University Science, 1989.
% \end{thebibliography}
% \vspace{12pt}
\end{document}